%% file: iclr2027_conference.tex
\documentclass{article}
\usepackage[utf8]{inputenc}
\usepackage{graphicx}
\usepackage{iclr2027_conference,times}

\input{math_commands.tex}

\usepackage{hyperref}
\usepackage{url}
\usepackage{booktabs}
\usepackage{multirow}
\usepackage{bbding}
\usepackage{pifont}

\title{Decoupling Spherical Reasoning from Dense Prediction for 360° Depth Estimation}

\author{Zhijie Shen$^1$, Chunyu Lin$^1$, Shuai Zheng$^1$, Feng Li$^2$, Runmin Cong$^3$, Huihui Bai$^1$ \& Yao Zhao$^1$\\
$^1$Beijing Jiaotong University\\
$^2$Hefei University of Technology \\
$^3$Shandong University 
}

\iclrfinalcopy
\begin{document}

\maketitle
\begin{abstract}
The equirectangular projection (ERP) is widely used for panoramic depth estimation, but its spatially varying distortion makes geometry-consistent feature modeling challenging. We revisit panoramic depth estimation by decoupling contextual modeling in native spherical space from dense ERP prediction. To this end, we propose a Fibonacci Spherical Graph (FSG) as an intermediate reasoning space to lift ERP features onto quasi-uniform Fibonacci nodes on the sphere and capture local and long-range dependencies through complementary spherical neighborhoods. The resulting spherical discretization distributes graph nodes approximately uniformly over the spherical surface, reducing the over-representation of highly stretched regions during relational modeling. Operating on a compact set of Fibonacci nodes also avoids the computational burden of constructing and processing a graph at full ERP resolution. To bridge spherical reasoning and dense prediction, we propose a Spherical Context Conditioning (SCC) module that adaptively modulates dense ERP features with the enhanced spherical representation, allowing spherical context to guide pixel-aligned depth prediction. Extensive experiments on three benchmarks demonstrate that the proposed method consistently achieves superior depth accuracy over existing approaches.
\end{abstract}

\section{Introduction}

Panoramic images capture a complete $360^\circ$ observation of the surrounding environment and provide rich spatial information for scene understanding, robotic navigation, augmented reality, and 3D reconstruction~\citep{chang2017matterport3d,albanis2021pano3d}. Monocular panoramic depth estimation aims to recover dense scene depth from a single panoramic observation and serves as a fundamental task for understanding surrounding 3D environments. Unlike perspective images, panoramic images represent a spherical observation of a scene, where spatial relationships are naturally defined on the sphere rather than on a planar image domain.

To process panoramic images with existing vision architectures, most methods adopt the equirectangular projection (ERP), which unfolds the spherical image into a regular planar representation and enables the direct application of mature dense prediction networks~\citep{zioulis2018omnidepth,wang2020bifuse}. However, this spherical-to-planar transformation inevitably alters the spatial organization of the scene. Due to the latitude-dependent distortion of ERP, image-space distances and neighborhood structures undergo different degrees of stretching across the panorama~\citep{tateno2018distortion,chen2021distortion}. Consequently, pixel neighborhoods defined on the ERP grid do not uniformly reflect spatial relationships on the sphere. A local window of identical size can correspond to different spherical extents and geometric structures at different locations, causing feature aggregation to capture inconsistent scene contexts. As a result, models operating directly on ERP representations may learn contextual dependencies influenced by the projection space rather than the spherical geometry of the scene, making geometry-consistent feature modeling challenging for panoramic depth estimation.

Existing panoramic depth estimation methods mainly alleviate these challenges through projection-aware feature learning. ERP-based approaches adapt convolution operations by modifying sampling locations or receptive fields according to spatially varying distortion in the equirectangular domain~\citep{tateno2018distortion,coors2018spherenet,khasanova2019geometry}. Alternatively, multi-projection approaches introduce additional representations, such as cubemap~\citep{wang2020bifuse,Wang2022BiFuseSA,shen2026revisiting} or tangent views~\citep{Li2022OmniFusion3M,Shen2022PanoFormerPT}, to complement ERP features and reduce the impact of severe projection distortion~\citep{wang2020bifuse,jiang2021unifuse,Li2022OmniFusion3M,Ai2023HRDFuseM3,Ai_2024_CVPR}. Recent methods further incorporate attention mechanisms, geometric constraints, and cross-projection fusion to enhance contextual modeling~\citep{Shen2022PanoFormerPT,Ai2023HRDFuseM3,Yun_2023_ICCV,Li2023mathcalA}. Although these approaches have achieved significant progress, feature interactions in existing panoramic depth estimation pipelines are still mainly performed in equirectangular or other planar projection spaces. Consequently, the contextual relationships learned by these methods remain constrained by the spatial organization of the projection space, rather than being directly established according to spherical geometry.

This observation motivates us to reconsider how feature relationships are modeled in panoramic depth estimation. Rather than continuously adapting planar representations to compensate for projection distortion, we aim to construct a geometry-consistent spherical reasoning space where feature interactions are established according to the geometric structure of the sphere. Since the spherical surface does not naturally follow a regular planar grid~\citep{cohen2018spherical,jiang2019spherical,lee2019spherephd}, such a reasoning space requires a representation that can flexibly describe spherical neighborhoods and relational interactions. This provides more appropriate support for contextual modeling by allowing relationships among features to be aggregated according to their geometric configurations on the sphere. Importantly, the spherical reasoning space is designed to complement rather than replace ERP representations, since ERP remains effective for dense pixel-aligned prediction.

Based on this principle, we introduce a geometry-consistent spherical reasoning space for monocular panoramic depth estimation. Instead of performing all feature interactions in the distorted ERP domain or replacing ERP prediction with spherical decoding, we decouple contextual reasoning from dense prediction.
Specifically, we instantiate this reasoning space with a Fibonacci Spherical Graph (FSG), where spherical relationships are modeled independently from the pixel-aligned ERP representation. Instead of defining contextual interactions solely on projected image grids, FSG establishes feature relationships directly over spherical geometry, allowing spatial dependencies to be modeled independently of the latitude-dependent deformation of ERP. Fibonacci discretization provides a compact and approximately uniform support for this reasoning process, making spherical relational modeling practical without replacing the dense ERP representation.

To bridge spherical reasoning and dense prediction, we further introduce a Spherical Context Conditioning (SCC) module.
SCC transfers geometry-aware spherical context back to ERP features while preserving the dense spatial correspondence required for depth estimation. Hence, FSG reasons over scene geometry on the sphere, while ERP features remain responsible for spatially aligned depth estimation.

Extensive experiments on three widely used panoramic depth benchmarks demonstrate that the proposed method consistently improves depth estimation accuracy over existing approaches. Ablation studies further verify the effectiveness of spherical relational modeling and spherical context conditioning. Our main contributions are summarized as follows:

\begin{itemize}
\item We introduce a new formulation for panoramic depth estimation that separates geometry-aware spherical reasoning from dense ERP prediction.

\item We propose a Fibonacci Spherical Graph (FSG) as a compact intermediate reasoning space, where contextual dependencies are modeled according to spherical geometry rather than ERP sampling topology.

\item We design a Spherical Context Conditioning (SCC) module to effectively transfer spherical context into pixel-aligned ERP features for dense depth estimation.

\end{itemize}

\section{Related Work}
\paragraph{Monocular panoramic depth estimation.}
Most monocular panoramic depth estimation methods adopt equirectangular projection (ERP) because it preserves a dense rectangular layout and is compatible with standard CNN- and Transformer-based architectures~\citep{zioulis2018omnidepth}. However, the latitude-dependent distortion of ERP causes inconsistent spatial relationships across the panorama. Early works address this issue through distortion-aware convolutions, adaptive sampling, and geometry-aware receptive fields~\citep{tateno2018distortion,chen2021distortion,su2019kernel}, but feature interactions are still defined on the distorted ERP grid.

Another line of research combines ERP with complementary projections. BiFuse and UniFuse fuse ERP and cubemap features~\citep{wang2020bifuse,jiang2021unifuse}, while OmniFusion, HRDFuse, and PanoFormer exploit tangent or multi-projection representations to enhance contextual modeling~\citep{Li2022OmniFusion3M,Ai2023HRDFuseM3,Shen2022PanoFormerPT}. Although these approaches improve distortion handling and global context, feature relationships are still established within individual projection domains or through cross-projection fusion, rather than being explicitly organized according to spherical geometry.

Recent methods further explore spherical representations for panoramic depth estimation. SphereDepth performs feature extraction and prediction on spherical meshes~\citep{yan2022spheredepth}, $\mathcal{S}^{2}$Net employs equal-area HEALPix grids for spherical decoding~\citep{Li2023mathcalA}, and Elite360D combines ERP features with an icosahedral point representation through bi-projection fusion~\citep{Ai_2024_CVPR}. These methods demonstrate the effectiveness of spherical geometry, but the sphere is mainly explored as a prediction domain or an auxiliary representation. In contrast, we introduce a compact spherical reasoning space where feature relationships are modeled according to spherical geometry while dense depth prediction remains in ERP space.

\paragraph{Spherical representation and geometric modeling.}
Spherical representations have been widely studied for omnidirectional vision. Spherical CNNs and equivariant networks define operators directly on spherical domains~\citep{cohen2018spherical,esteves2018learning}, while SphereNet and related approaches adapt convolution operations to address ERP distortion~\citep{coors2018spherenet,lee2019spherephd,jiang2019spherical}. These studies mainly focus on constructing geometry-aware feature representations for spherical signals.

Graph-based spherical models provide a flexible formulation for irregular spherical structures. DeepSphere represents the sphere as a graph and performs graph convolution over spherical neighborhoods~\citep{Defferrard2020DeepSphereAG}. Recent works have also explored Fibonacci-based spherical discretizations for efficient spherical representations, such as organizing spherical Gaussian primitives for high-resolution panorama synthesis~\citep{zhang2025pansplat}. 

Different from these approaches, FSG employs Fibonacci nodes as a compact spherical reasoning space rather than a final representation or prediction domain. The proposed design introduces spherical geometry-aware relational modeling while retaining ERP for dense pixel-aligned prediction.

%
%

\section{Methodology}
\label{sec:methodology}

\subsection{Overview}
\label{sec:method_overview}

Panoramic depth estimation requires contextual reasoning across the spherical field of view while preserving the pixel correspondence needed for dense prediction. ERP offers a convenient representation for the latter, but its latitude-dependent distortion makes image-grid neighborhoods an uneven basis for the former. We therefore assign these roles to complementary representations: a Fibonacci Spherical Graph (FSG) organizes contextual interaction on the sphere, and Spherical Context Conditioning (SCC) transfers the resulting context to dense ERP features.

\begin{figure}[t]
\centering
\includegraphics[width=\linewidth]{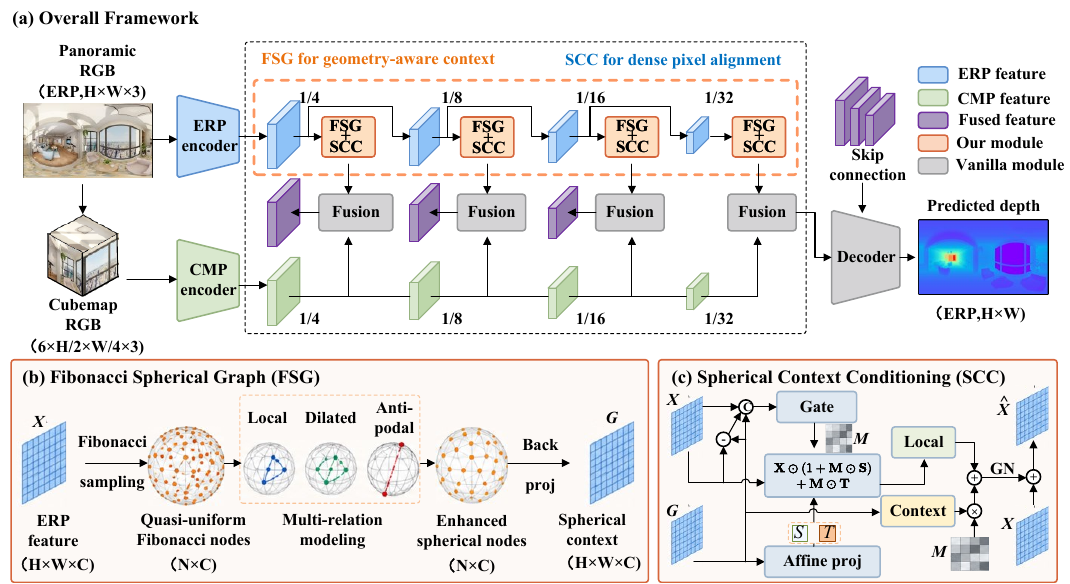}
\caption{
Overview of the proposed framework.
(a) FSG and SCC refine multi-scale ERP features before ERP--cubemap fusion.
(b) FSG samples features onto Fibonacci nodes, aggregates local, dilated,
and antipodal context, and back-projects the enhanced features to ERP.
(c) SCC uses spherical context $G$ to condition dense ERP features $X$
through gated modulation and context injection.
The schematic is elaborated by the operators and residual update
in Sec.~\ref{sec:scc}.
}
\label{fig:framework}
\end{figure}

Figure~\ref{fig:framework}(a) shows how this separation is incorporated into UniFuse~\citep{jiang2021unifuse}. We retain its dual-projection encoders, fusion modules, and ERP decoder, and independently refine the ERP features at four scales, from $1/4$ to $1/32$ resolution, before their corresponding fusion stages. The FSG branch in Figure~\ref{fig:framework}(b) maps an ERP feature $X_l$ to spherical context $G_l$, which the SCC module in Figure~\ref{fig:framework}(c) uses to produce the refined feature $\hat X_l$. For a feature map $X_l\in\mathbb R^{B\times C_l\times H_l\times W_l}$, this process is
\begin{equation}
\label{eq:fsg_pipeline}
\begin{aligned}
V_l&=\mathcal S_l(X_l,\mathcal P_l),
&Z_l&=\mathcal R_l(V_l;\mathcal P_l),\\
G_l&=\mathcal B_l(Z_l,\mathcal P_l),
&\hat X_l&=X_l+\mathcal C_l(X_l,G_l).
\end{aligned}
\end{equation}
Here, $\mathcal S_l$ samples features onto spherical nodes $\mathcal P_l$, $\mathcal R_l$ models their relationships, $\mathcal B_l$ returns the enhanced context to ERP, and $\mathcal C_l$ produces a residual update. The sphere thus serves as an intermediate reasoning space within the dense prediction pipeline. We omit the scale index below.

\subsection{Fibonacci Spherical Representation}
\label{sec:fibonacci_support}

The first consideration is how to distribute the support for spherical reasoning. Uniformly spaced ERP pixels represent progressively smaller spherical areas toward the poles. Using these locations as graph nodes therefore allocates a disproportionate share of the node budget to high-latitude regions. We use $N$ Fibonacci nodes $\mathcal P=\{\mathbf p_i\}_{i=0}^{N-1}$ on the unit sphere, combining uniformly spaced vertical coordinates with golden-angle increments to obtain approximately uniform spherical coverage.

Following the sampling stage in Figure~\ref{fig:framework}(b), each node gathers an ERP feature by bilinear sampling at its corresponding viewing direction, producing $V\in\mathbb R^{B\times N\times C}$. Horizontal circular padding connects samples across the ERP seam. This construction gives relational modeling a more balanced distribution of viewing directions while allowing its node count to be chosen independently of ERP resolution. The original dense features remain available for subsequent prediction, and gradients propagate through sampling to the encoder. Node coordinates are fixed; their construction and sampling mappings are provided in Appendix~\ref{app:fibonacci_sampling}.

\subsection{Multi-Relation Spherical Graph Reasoning}
\label{sec:spherical_graph}

A balanced node distribution provides the support for reasoning, but the connections determine which evidence can be brought together. Nearby viewing directions provide local context, while larger scene structures extend across broader angular ranges. We therefore construct the local, dilated, and antipodal relations illustrated in Figure~\ref{fig:framework}(b) over the same nodes. The local relation selects the $k$ nearest nodes, including the center node. The dilated relation selects the next $k$ nodes in the distance ordering, extending interaction beyond the immediate neighborhood. The antipodal relation selects the $k$ nodes nearest to $-\mathbf p_i$, providing direct communication with the opposite part of the panorama.

For relation $r$, let $\mathcal N_r(i)$ denote the neighbors of node $i$. Given input node features $h_i$, relation-specific aggregation is
\begin{equation}
\label{eq:fsg_messages}
h_i^r=\phi_r\!\left(
\sum_{j\in\mathcal N_r(i)}
\alpha_{ij}^r
\bigl(W_rh_j+E_r\mathbf g_{ij}\bigr)
\right),
\end{equation}
where $W_r$ and $E_r$ project features and relative geometry, and $\mathbf g_{ij}$ encodes node displacement and angular separation. The operator $\phi_r$ applies the post-aggregation transformation. Affinities $\alpha_{ij}^r$ are fixed softmax-normalized cosine similarities, measured from $\mathbf p_i$ for local and dilated relations and from $-\mathbf p_i$ for the antipodal relation. To let the contribution of each angular range depend on the observed features, we combine the relation responses using node-wise weights:
\begin{equation}
\label{eq:fsg_relations}
\beta_i^r=\operatorname{softmax}_{r}
\bigl(\mathbf a_r^\top h_i^r+b_r\bigr),
\qquad
u_i=\sum_r\beta_i^r h_i^r.
\end{equation}

Spherical geometry determines which directions exchange information and how messages are aggregated within each relation; learned relation weights adapt their combination at each node. This allows nearby evidence and distant context to contribute differently across the panorama. Residual graph blocks with feed-forward refinement produce the enhanced representation $Z$. Complete weight and block definitions are given in Appendix~\ref{app:graph_details}.

\subsection{Spherical Context Conditioning}
\label{sec:scc}

The remaining challenge is to make spherical reasoning useful for dense prediction without making pixel-level detail depend solely on the graph representation. The back-projection stage in Figure~\ref{fig:framework}(b) interpolates enhanced node features according to each ERP pixel's viewing direction, obtaining a context map $G\in\mathbb R^{B\times C\times H\times W}$. This restores spatial correspondence, but interpolation aggregates node features and does not guarantee recovery of the original dense feature variations. SCC, illustrated in Figure~\ref{fig:framework}(c), therefore keeps the ERP feature $X$ as the carrier of dense information and uses $G$ to condition its update. The interpolation rule is provided in Appendix~\ref{app:backprojection}.

The affine projection in Figure~\ref{fig:framework}(c) predicts a bounded scale adjustment $S$ and an additive shift $T$ from spherical context. The Gate branch jointly uses the ERP feature, spherical context, and their difference to control modulation at each spatial location and channel. Making the normalization operations explicit, let $\bar X$ and $\bar G$ denote separately group-normalized features. The conditioning update is
\begin{equation}
\label{eq:scc_update}
\begin{aligned}
\left[S_0,T\right]&=f_{\mathrm{aff}}(\bar G),
& S&=a\tanh(S_0),\\
M&=\sigma\!\left(
f_g([\bar X,\bar G,\bar X-\bar G])
\right),\\
X_c&=X\odot(1+M\odot S)+M\odot T,\\
\mathcal C(X,G)&=\mathcal N_o\!\left(
f_{\mathrm{local}}(X_c)
+M\odot f_{\mathrm{ctx}}(\bar G)
\right).
\end{aligned}
\end{equation}

Here, $a=0.5$, $\sigma$ is sigmoid, brackets denote channel concatenation, and $\odot$ denotes element-wise multiplication. The Local branch processes $X_c$ through a depth-wise convolution with horizontal circular padding. The Context branch projects $\bar G$, and its response is multiplied by the same gate $M$. GroupNorm $\mathcal N_o$ normalizes the combined update, which is added to the original $X$ as specified in Eq.~\ref{eq:fsg_pipeline}. This coupling lets spherical relationships influence both the transformation of existing ERP features and the context added to them, while retaining a direct dense feature pathway. Operator details and module configurations are given in Appendices~\ref{app:scc_details} and~\ref{app:module_config}.

\subsection{Objective Function}
\label{sec:training_objective}

Following previous works~\citep{shen2026revisiting,lee2025hush}, our objective function consists of two terms. We use the BerHu loss $\mathcal{L}_{\mathrm{BerHu}}$ for depth regression. Given the predicted depth $\hat D$, ground-truth depth $D$, and the set of valid pixels $\Omega_Q$, it is defined as
\begin{equation}
\label{eq:berhu_loss}
\mathcal{L}_{\mathrm{BerHu}}
=
\frac{1}{|\Omega_Q|}
\sum_{p\in\Omega_Q}
\begin{cases}
e_p, & e_p \leq c,\\[2pt]
\dfrac{e_p^2+c^2}{2c}, & e_p > c,
\end{cases}
\qquad
e_p = |\hat D_p-D_p|,
\end{equation}
where $c=0.2$ denotes the BerHu transition threshold.

We further adopt the spherical gradient loss $\mathcal{L}_{\mathrm{sg}}$ introduced in PGFuse~\citep{shen2026revisiting} to supervise spatial depth variations,
\begin{equation}
\label{eq:sg_loss}
\mathcal{L}_{\mathrm{sg}}
=
\frac{1}{|\Omega_Q^{\nabla}|}
\sum_{p\in\Omega_Q^{\nabla}}
\left\|
\nabla_{s}\hat D_p-\nabla_{s}D_p
\right\|_1,
\end{equation}
where $\nabla_s$ denotes the spherical gradient operator and
$\Omega_Q^{\nabla}$ contains valid gradient pairs.

The overall objective is formulated as
\begin{equation}
\label{eq:training_loss}
\mathcal{L}
=
\mathcal{L}_{\mathrm{BerHu}}
+
\lambda_{\mathrm{sg}}\mathcal{L}_{\mathrm{sg}},
\end{equation}
where $\lambda_{\mathrm{sg}}$ is set to $0.5$ following PGFuse~\citep{shen2026revisiting} and HUSH~\citep{lee2025hush}.

\begin{table}[t]
\centering
\small
\caption{
Quantitative comparison with existing panoramic depth estimation methods. For fair comparison, $^{*}$ denotes re-evaluation under the Elite360D protocol, while $^{\dagger}$ denotes retraining with a ResNet-34 backbone. All remaining baseline results are taken from PGFuse~\citep{shen2026revisiting}. Best and second-best results are shown in bold and underlined, respectively.
}
\label{tab:sota}
\setlength{\tabcolsep}{6.0pt}
\begin{tabular}{llcccccc}
\toprule
Dataset & Method
& Abs Rel $\downarrow$
& Sq Rel $\downarrow$
& RMSE $\downarrow$
& $\delta_1(\%)$ $\uparrow$
& $\delta_2(\%)$ $\uparrow$
& $\delta_3(\%)$ $\uparrow$ \\
\midrule

\multirow{11}{*}{M3D}
& EGFormer
& 0.1473 & 0.1517 & 0.6025 & 81.58 & 93.90 & 97.35 \\
& PanoFormer
& 0.1051 & 0.0966 & 0.4929 & 89.08 & 96.23 & 98.31 \\
& BiFuse
& 0.1126 & 0.0992 & 0.5027 & 88.00 & 96.13 & 98.47 \\
& BiFuse++
& 0.1123 & 0.0915 & 0.4853 & 88.12 & 96.56 & 98.69 \\
& UniFuse
& 0.1144 & 0.0936 & 0.4835 & 87.85 & 96.59 & 98.73 \\
& OmniFusion
& 0.1161 & 0.1007 & 0.4931 & 87.72 & 96.15 & 98.44 \\
& HRDFuse
& 0.1172 & 0.0971 & 0.5025 & 86.74 & 96.17 & 98.49 \\
& Elite360D
& 0.1115 & 0.0914 & 0.4875 & 88.15 & 96.46
& \underline{98.74} \\
& PGFuse
& 0.1067 & 0.0858 & 0.4624 & 89.06 & 96.75 & 98.72 \\
& HUSH\textdagger
& \underline{0.0957}
& \underline{0.0798}
& \underline{0.4556}
& \underline{90.91}
& \underline{96.83}
& 98.66 \\
& Ours
& \textbf{0.0949}
& \textbf{0.0776}
& \textbf{0.4455}
& \textbf{91.16}
& \textbf{97.01}
& \textbf{98.87} \\
\midrule

\multirow{8}{*}{S2D3D}
& EGFormer
& 0.1528 & 0.1408 & 0.4974 & 81.85 & 93.38 & 97.36 \\
& PanoFormer
& 0.1122 & 0.0786 & 0.3945 & 88.74 & 95.84 & 98.59 \\
& OmniFusion
& 0.1154 & 0.0775 & 0.3809 & 86.74 & 96.03 & 98.71 \\
& UniFuse
& 0.1124 & 0.0709 & 0.3555 & 87.06 & 97.04 & 98.99 \\
& Elite360D
& 0.1182 & 0.0728 & 0.3756 & 88.72 & 96.84 & 98.92 \\
& PGFuse*
& 0.1035 & 0.0645 & 0.3680 & 88.98 & 96.79 & 99.00 \\
& HUSH\textdagger
& \underline{0.0982}
& \underline{0.0587}
& \underline{0.3446}
& \underline{89.15}
& \underline{97.46}
& \underline{99.12} \\
& Ours
& \textbf{0.0949}
& \textbf{0.0568}
& \textbf{0.3420}
& \textbf{89.53}
& \textbf{97.68}
& \textbf{99.18} \\
\midrule

\multirow{8}{*}{S3D}
& EGFormer
& 0.2205 & 0.4509 & 0.6841 & 79.79 & 90.71 & 94.55 \\
& PanoFormer
& 0.2549 & 0.4949 & 0.7937 & 74.70 & 89.15 & 93.97 \\
& BiFuse
& 0.1573 & 0.2455 & 0.5213 & 85.91 & 94.00 & 96.72 \\
& UniFuse
& 0.1506 & 0.2319 & 0.5016 & 85.42 & 93.99 & 96.76 \\
& Elite360D
& 0.1480 & \underline{0.2215} & 0.4961
& 87.41 & 94.34 & 96.66 \\
& PGFuse
& \underline{0.1401}
& 0.2499
& \underline{0.4394}
& 87.78
& 94.49
& \underline{96.82} \\
& HUSH\textdagger
& 0.1427 & 0.2489 & 0.4571
& \underline{88.45}
& \underline{94.56}
& 96.71 \\
& Ours
& \textbf{0.1289}
& \textbf{0.2129}
& \textbf{0.4046}
& \textbf{89.12}
& \textbf{95.25}
& \textbf{97.09} \\
\bottomrule
\end{tabular}
\end{table}

\section{Experiments}
\label{sec:experiments}

\subsection{Experimental Setup}
\label{subsec:setup}

\paragraph{Datasets.}
We evaluate on Matterport3D~\citep{chang2017matterport3d}, Stanford2D3D~\citep{armeni2017joint}, and Structured3D~\citep{Zheng2019Structured3DAL}. The first two datasets contain real indoor panoramas, while Structured3D provides synthetic indoor scenes with depth annotations. Together, they allow us to assess the method on both captured and rendered panoramic observations. We follow the training and evaluation protocols adopted by previous methods~\citep{jiang2021unifuse,Ai_2024_CVPR} for each benchmark.

\paragraph{Evaluation metrics.}
We report Absolute Relative Error (Abs Rel), Squared Relative Error (Sq Rel), Root Mean Squared Error (RMSE), and threshold accuracies $\delta_t$, $t\in\{1,2,3\}$, over valid depth pixels. The latter measure the percentage of pixels satisfying $\max(\hat D/D,D/\hat D)<1.25^t$. RMSE emphasizes larger absolute depth errors, whereas the threshold accuracies measure how frequently predictions meet specified relative-error bounds.

\paragraph{Implementation details.}
We implement our approach using the PyTorch framework and conduct all experiments on a single NVIDIA GeForce RTX 3090 GPU. We adopt a ResNet-34 backbone initialized with ImageNet-1K pretrained weights. The model is trained using the Adam optimizer with an initial learning rate of $1\times10^{-4}$, and a batch size of 8. Other optimizer hyperparameters follow the default settings. During training, we employ horizontal flipping, horizontal circular rotation, and luminance augmentation following previous panoramic depth estimation methods~\citep{wang2020bifuse,jiang2021unifuse}. The node counts, neighborhood sizes, and graph depths are specified in Appendix~\ref{app:module_config}; sampling, propagation, and conditioning details are given in Appendices~\ref{app:fibonacci_sampling}--\ref{app:scc_details}.

\subsection{Comparison with State-of-the-Art Methods}
\label{subsec:sota}

\paragraph{Quantitative comparison.}
Table~\ref{tab:sota} compares our method with existing panoramic depth estimation approaches, including EGFormer~\citep{Yun_2023_ICCV}, PanoFormer~\citep{Shen2022PanoFormerPT}, BiFuse~\citep{wang2020bifuse}, BiFuse++~\citep{Wang2022BiFuseSA}, UniFuse~\citep{jiang2021unifuse}, OmniFusion~\citep{Li2022OmniFusion3M}, HRDFuse~\citep{Ai2023HRDFuseM3}, Elite360D~\citep{Ai_2024_CVPR}, PGFuse~\citep{shen2026revisiting}, and HUSH~\citep{lee2025hush}. Our method achieves the lowest errors and highest threshold accuracies among the compared methods on all three datasets.


Table~\ref{tab:sota} shows that our method consistently outperforms existing approaches on all three benchmarks, with strong performance in both RMSE and $\delta$ metrics. This indicates that the method reduces large depth errors while maintaining reliable prediction across the panorama. The improvement is closely related to our decoupled design, where FSG performs context modeling in the native spherical space to alleviate ERP distortion and sampling imbalance, while SCC transfers the resulting spherical context back to the dense ERP stream for pixel-aligned depth prediction.

\begin{figure}[t]
\centering
\includegraphics[width=\linewidth]{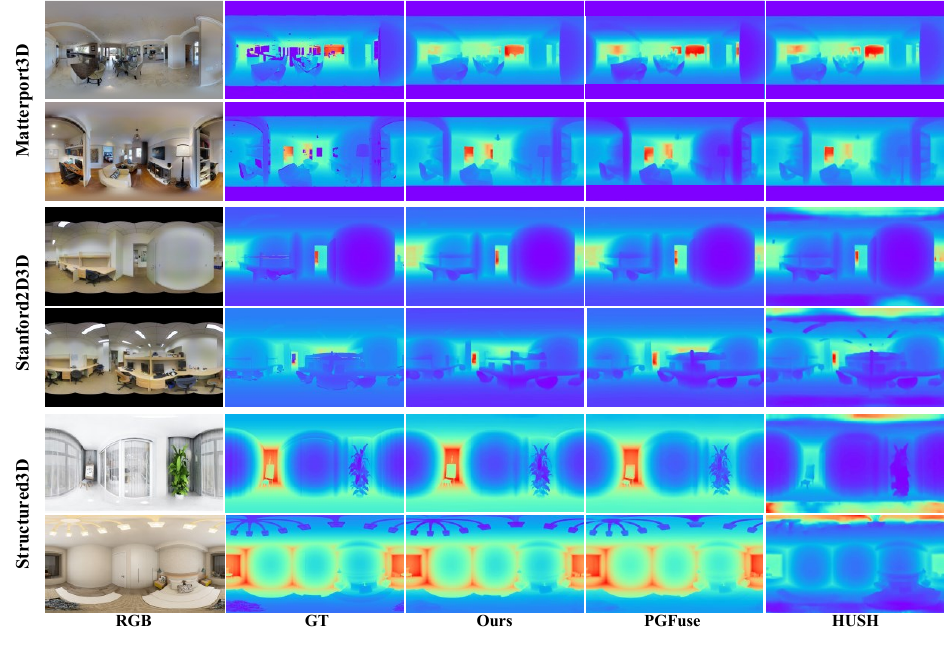}
\caption{
Qualitative comparison results on Matterport3D, Stanford2D3D,
and Structured3D. Best viewed in color.
}
\label{fig:qualitative}
\end{figure}

\paragraph{Qualitative comparison.}
Figure~\ref{fig:qualitative} presents qualitative comparisons on all three benchmarks. On Matterport3D, our predictions follow the ground-truth depth variation more consistently over extended walls and floor regions while preserving clear transitions around openings and furniture. On Stanford2D3D, large surfaces exhibit more coherent depth variation, and distant structures remain better aligned with the ground truth. Similar behavior is observed on Structured3D, where our method better preserves room-scale geometry and the spatial organization of extended surfaces.

Local depth transitions around object and structural boundaries are also well preserved rather than being over-smoothed by broader contextual aggregation. These visual results complement the quantitative evaluation by showing that the proposed method improves large-scale structural consistency while retaining local geometric detail.

\subsection{Ablation Studies and Analysis}
\label{subsec:ablation}

We conduct ablations on Matterport3D to examine the contributions of graph reasoning, SCC, and spherical node support. Table~\ref{tab:component_ablation} compares five controlled variants. The first row corresponds to the UniFuse baseline retrained with the spherical gradient loss. To isolate graph reasoning without SCC, we back-project the graph-enhanced Fibonacci features to the ERP space and add them element-wise to the original ERP features (the second row). Then the sampled Fibonacci features are directly back-projected without graph reasoning and incorporated into the ERP stream through SCC (the third row). We then replace Fibonacci nodes with uniformly sampled ERP-grid nodes while retaining graph reasoning and SCC (the fourth row). The complete model combines Fibonacci-based graph reasoning with SCC (the fifth row). The ERP-grid and Fibonacci variants use matched node and edge counts, allowing the influence of node support to be examined independently.

\paragraph{Effect of spherical node support.}
From Table~\ref{tab:component_ablation}, we can observe that replacing ERP-grid nodes with Fibonacci nodes reduces RMSE, while improving $\delta_1$. Since the remaining components are unchanged, this comparison directly reflects the influence of node support. Fibonacci nodes distribute the representation budget more uniformly over the native spherical space, reducing the latitude-dependent redundancy introduced by ERP-grid sampling.


\paragraph{Effect of graph reasoning.}
Adding graph reasoning to the Fibonacci representation improves both RMSE and $\delta_1$. Fibonacci sampling determines how viewing directions are distributed over the sphere, whereas graph reasoning explicitly establishes dependencies among them. The additional gain shows that a balanced spherical support alone is insufficient and that contextual interaction over the spherical topology contributes further to depth estimation.

\begin{table}[t]
\centering
\caption{
Ablation of node support, graph reasoning, and SCC on Matterport3D.
ERP Grid uses uniformly sampled ERP nodes with matched node and edge counts.
}
\label{tab:component_ablation}
\setlength{\tabcolsep}{6pt}
\begin{tabular}{lccccc}
\toprule
Support & Graph & SCC
& RMSE $\downarrow$
& Abs Rel $\downarrow$
& $\delta_1$ $\uparrow$ \\
\midrule
ERP
& -- & --
& 0.4737 & 0.1036 & 88.36 \\
Fibonacci
& \checkmark & --
& 0.4617 & 0.0968 & 90.62 \\
Fibonacci
& -- & \checkmark
& 0.4577 & 0.1017 & 90.01 \\
ERP Grid
& \checkmark & \checkmark
& 0.4558 & 0.0999 & 90.29 \\
Fibonacci
& \checkmark & \checkmark
& \textbf{0.4455}
& \textbf{0.0949}
& \textbf{91.16} \\
\bottomrule
\end{tabular}
\end{table}

\paragraph{Effect of SCC.}
As shown in Table~\ref{tab:component_ablation}, SCC consistently improves all evaluation metrics over direct residual fusion. This suggests that simply injecting the back-projected spherical features into the ERP stream is insufficient. By adaptively conditioning the dense ERP features with graph-derived spherical context, SCC enables more selective information transfer and better preserves spatially localized depth structures.

\begin{table}[ht]
\centering
\caption{
Ablation of spherical relations on Matterport3D. Fibonacci support and SCC are retained.
}
\label{tab:relation_ablation}
\setlength{\tabcolsep}{5pt}
\begin{tabular}{lccc}
\toprule
Relations
& RMSE $\downarrow$
& Abs Rel $\downarrow$
& $\delta_1$ $\uparrow$ \\
\midrule
Local
& 0.4577 & 0.0994 & 90.42 \\
Local + Dilated
& 0.4615 & 0.1006 & 90.34 \\
Local + Antipodal
& 0.4501 & 0.0957 & 90.96 \\
Local + Dilated + Antipodal
& \textbf{0.4455}
& \textbf{0.0949}
& \textbf{91.16} \\
\bottomrule
\end{tabular}
\end{table}

\paragraph{Effect of spherical relation configurations.}
Table~\ref{tab:relation_ablation} shows that adding antipodal connections clearly improves the local graph, whereas adding the dilated relation to the local graph does not, indicating that simply enlarging the interaction range is insufficient. The best performance is achieved by combining all three relations, suggesting that local, intermediate-range, and long-range interactions provide complementary structural context. Adaptive relation fusion further integrates these context paths for effective multi-range spherical reasoning.

\paragraph{Effect across latitude regions.}
Table~\ref{tab:latitude_ablation} further compares ERP-grid and Fibonacci support across different latitude regions. Fibonacci support improves all three regions, with the largest gain observed in the equatorial region. Under uniform ERP-grid sampling, a fixed node budget is distributed uniformly in image coordinates even though equal latitude intervals correspond to unequal spherical surface areas. This results in relatively redundant support toward high latitudes and comparatively insufficient representation of the lower-distortion equatorial region. Redistributing the nodes more uniformly over the sphere alleviates this imbalance and assigns more effective representation capacity to regions with larger spherical coverage and less projection distortion. The more pronounced improvement near the equator is consistent with this motivation, while the gains across all latitude regions indicate that the benefit is not confined to a particular part of the panorama.

\begin{table}[h]
\centering
\caption{
Latitude-wise comparison of ERP-grid and Fibonacci node support on Matterport3D.
Both variants retain the same graph reasoning and SCC configuration.
}
\label{tab:latitude_ablation}
\setlength{\tabcolsep}{7pt}
\begin{tabular}{llccc}
\toprule
Support & Region
& RMSE $\downarrow$
& Abs Rel $\downarrow$
& $\delta_1$ $\uparrow$ \\
\midrule
ERP Grid
& Equatorial ($30^\circ\mathrm{S}$--$30^\circ\mathrm{N}$)
& 0.5736 & 0.1151 & 87.60 \\
Fibonacci
& Equatorial ($30^\circ\mathrm{S}$--$30^\circ\mathrm{N}$)
& \textbf{0.5585} & \textbf{0.1073} & \textbf{88.99} \\
ERP Grid
& Mid-latitude ($30^\circ\mathrm{S}$--$60^\circ\mathrm{S}$, $30^\circ\mathrm{N}$--$60^\circ\mathrm{N}$)
& 0.3219 & 0.0894 & 92.19 \\
Fibonacci
& Mid-latitude ($30^\circ\mathrm{S}$--$60^\circ\mathrm{S}$, $30^\circ\mathrm{N}$--$60^\circ\mathrm{N}$)
& \textbf{0.3151} & \textbf{0.0861} & \textbf{92.70} \\
ERP Grid
& High-latitude ($60^\circ\mathrm{S}$--$90^\circ\mathrm{S}$, $60^\circ\mathrm{N}$--$90^\circ\mathrm{N}$)
& 0.2676 & 0.0822 & 93.29 \\
Fibonacci
& High-latitude ($60^\circ\mathrm{S}$--$90^\circ\mathrm{S}$, $60^\circ\mathrm{N}$--$90^\circ\mathrm{N}$)
& \textbf{0.2605} & \textbf{0.0817} & \textbf{93.48} \\
\bottomrule
\end{tabular}
\end{table}

\paragraph{Computational complexity analysis.}
Table~\ref{tab:efficiency} shows that our method introduces moderate computational overhead over UniFuse. However, compared with PGFuse and HUSH, which achieve comparable depth estimation accuracy, our method requires fewer FLOPs and less training memory and achieves lower inference latency. Spherical reasoning is performed on a compact set of Fibonacci nodes rather than full-resolution ERP features, enabling low-cost contextual modeling.

\begin{table}[h]
\centering
\caption{
Computational comparison at an input resolution of $512\times1024$ with a batch size of $1$, where FLOPs are computed using \texttt{fvcore.nn.FlopCountAnalysis}.
}
\label{tab:efficiency}
\setlength{\tabcolsep}{7pt}
\begin{tabular}{lccc}
\toprule
Model
& FLOPs $\downarrow$
& Train Mem. (GB) $\downarrow$
& Inference latency $\downarrow$ \\
\midrule
UniFuse
& 96G & 2.96 & 21ms \\
PGFuse
& 126G & 3.63 & 41ms \\
HUSH
& 237G & 5.11 & 47ms \\
Ours
& 104G & 3.37 & 38ms \\
\bottomrule
\end{tabular}
\end{table}

\section{Conclusion}
\label{sec:conclusion}

We introduced a Fibonacci Spherical Graph as an intermediate reasoning space for monocular panoramic depth estimation. Quasi-uniform spherical nodes and multiple geometric relations organize contextual interaction, while SCC incorporates the resulting context into dense ERP features. This separates spherical relational modeling from pixel-aligned prediction within the existing dual-projection pipeline. Experiments on Matterport3D, Stanford2D3D, and Structured3D demonstrate improved accuracy over the compared methods. The controlled node-support comparison and component ablations support the benefits of spherical sampling and the joint use of graph reasoning and conditioning. These findings suggest that choosing a dedicated representation for contextual interaction is a promising strategy for incorporating panoramic geometry into dense prediction pipelines.


\subsection*{AI use statement}

In this work, we used generative AI tools for language editing and stylistic refinement. The experimental analyses, numerical results, and scientific conclusions were produced by the authors.

Additionally, generative AI tools were used to improve the presentation and clarity of author-written technical descriptions. All AI-assisted text was manually reviewed and revised to ensure consistency with the actual method, implementation, and experimental results. We take responsibility for the final content of this work, including text, claims or artifacts produced with the aid of generative AI.

\subsection*{Ethics statement}

This work focuses on monocular panoramic depth estimation and does not involve human-subject studies, collection of personal data, or the release of new datasets. All experiments are conducted on existing research datasets following their established evaluation protocols. We are not aware of specific ethical risks introduced by the proposed methodology beyond those generally associated with computer vision systems and their downstream deployment.

\subsection*{Reproducibility statement}

We provide the methodological details required to reproduce the proposed approach in the main paper, including the Fibonacci spherical representation, multi-relation graph construction, spherical context conditioning, and training objective. The experimental section specifies the datasets, evaluation metrics, implementation settings, and ablation configurations used in our evaluation. Additional architectural and implementation details are provided in the supplementary material. We will release the source code to facilitate reproduction of the reported results.

\bibliography{references}
\bibliographystyle{iclr2027_conference}
\clearpage

\appendix

\section{Additional Method Details}
\label{app:method_details}

\subsection{Fibonacci Nodes and ERP Sampling}
\label{app:fibonacci_sampling}

For $i=0,\ldots,N-1$, the fixed unit-sphere nodes are generated as
\begin{equation}
\label{eq:fibonacci_nodes}
\begin{aligned}
z_i&=1-\frac{2(i+1/2)}{N},
&\rho_i&=\sqrt{1-z_i^2},\\
\theta_i&=i\pi(3-\sqrt5),
&\mathbf p_i&=
[\rho_i\cos\theta_i,\rho_i\sin\theta_i,z_i]^\top.
\end{aligned}
\tag{A1}
\end{equation}

The half-step offset avoids placing nodes exactly at the poles. Let $\varphi_i=\arcsin(z_i)$ and $\lambda_i=\operatorname{atan2}(p_{i,y},p_{i,x})$. Under the coordinate convention used in our implementation, increasing $\varphi$ corresponds to increasing ERP row index. The pixel-center coordinates are
\begin{equation}
\label{eq:node_to_erp}
u_i=W\left[
\left(\frac{\lambda_i}{2\pi}+\frac12\right)\bmod1
\right]-\frac12,
\qquad
v_i=H\left(\frac{\varphi_i}{\pi}+\frac12\right)-\frac12.
\tag{A2}
\end{equation}

We circularly pad one column on each horizontal side before bilinear sampling, shifting $u_i$ by one in the padded tensor. Sampling uses \texttt{align\_corners=False} and border handling in the vertical direction. The sampling operation is differentiable with respect to the feature values; node coordinates are fixed.

\subsection{Relation Construction and Graph Blocks}
\label{app:graph_details}

For each node $i$, we sort all nodes by decreasing $\mathbf p_i^\top\mathbf p_j$. The local neighborhood contains the first $k$ entries, including the center node $i$, while the dilated neighborhood contains entries $k+1$ through $2k$. The antipodal neighborhood contains the $k$ nodes with the largest values of $-\mathbf p_i^\top\mathbf p_j$. Each node therefore receives $k$ messages from each relation. The resulting per-node neighbor lists are used directly without additional symmetrization.

The relative geometry of a propagation pair $(i,j)$ is encoded as
\begin{equation}
\label{eq:relative_geometry}
\mathbf g_{ij}=
\left[
\mathbf p_j-\mathbf p_i;
\arccos\!\left(
\operatorname{clip}(\mathbf p_i^\top\mathbf p_j,-1,1)
\right)
\right]
\in\mathbb R^4.
\tag{A3}
\end{equation}

Within each relation, the aggregation weights are determined solely by the spherical node coordinates,
\begin{equation}
\label{eq:geometric_affinities}
\alpha_{ij}^r=
\frac{\exp(s_{ij}^r/\tau_e)}
{\sum_{t\in\mathcal N_r(i)}\exp(s_{it}^r/\tau_e)},
\qquad
s_{ij}^r=
\begin{cases}
\mathbf p_i^\top\mathbf p_j,
&r\in\{\mathrm{local},\mathrm{dilated}\},\\
-\mathbf p_i^\top\mathbf p_j,
&r=\mathrm{antipodal},
\end{cases}
\tag{A4}
\end{equation}
where $\tau_e=0.2$. The displacement and angular separation in Eq.~\ref{eq:relative_geometry} are always computed from the actual propagation pair $(\mathbf p_i,\mathbf p_j)$, including antipodal edges. Neighbor indices, relative-geometry encodings, and aggregation weights $\alpha_{ij}^r$ are precomputed from the node coordinates and remain fixed throughout training.

In Eq.~\ref{eq:fsg_messages}, the post-aggregation operator is
$\phi_r(m)=\operatorname{LN}_r(\operatorname{PReLU}_r(m+c_r))$,
where $c_r$ is a learned bias. Each relation uses independent feature and geometric projections. The resulting relation-specific responses are further evaluated by independent learned relation-scoring functions to obtain the node-wise fusion weights $\beta_i^r$ in Eq.~\ref{eq:fsg_relations}. Thus, $\alpha_{ij}^r$ provides fixed geometric weighting within each relation, whereas $\beta_i^r$ performs feature-adaptive weighting across relations.

Writing the fused responses from Eq.~\ref{eq:fsg_relations} as a matrix $U$, each graph block updates its input $H$ as
\begin{equation}
\label{eq:graph_residual}
\tilde H=\operatorname{LN}(H+U),
\qquad
H^+=\tilde H+\operatorname{FFN}(\tilde H).
\tag{A5}
\end{equation}

The FFN consists of LayerNorm, a linear expansion from $C$ to $2C$, GELU, and a linear projection back to $C$. We use two graph blocks at each scale, with learned $C$-to-$C$ linear projections before the first block and after the second. Dropout is set to zero. The final projection produces the enhanced node representation $Z$ for back-projection to the ERP space.

\subsection{Spherical Back-Projection}
\label{app:backprojection}

For the ERP pixel at row $v$ and column $u$, its unit-sphere direction is
\begin{equation}
\label{eq:pixel_direction}
\begin{aligned}
\varphi_{uv}&=
\pi\left(\frac{v+1/2}{H}-\frac12\right),
&\lambda_{uv}&=
2\pi\left(\frac{u+1/2}{W}-\frac12\right),\\
\mathbf q_{uv}&=
[\cos\varphi_{uv}\cos\lambda_{uv},
\cos\varphi_{uv}\sin\lambda_{uv},
\sin\varphi_{uv}]^\top.
\end{aligned}
\tag{A6}
\end{equation}

We select the $k_b$ nodes with largest $\mathbf q_{uv}^\top\mathbf p_j$, denoted by $\mathcal N_b(u,v)$, and interpolate their enhanced features:
\begin{equation}
\label{eq:backprojection}
\begin{aligned}
w_{uv,j}&=
\frac{\exp(\mathbf q_{uv}^\top\mathbf p_j/\tau_b)}
{\sum_{t\in\mathcal N_b(u,v)}
\exp(\mathbf q_{uv}^\top\mathbf p_t/\tau_b)},\\
G_{uv}&=
\sum_{j\in\mathcal N_b(u,v)}w_{uv,j}Z_j,
\qquad \tau_b=0.05.
\end{aligned}
\tag{A7}
\end{equation}

The interpolation indices and weights depend only on the spherical node set and ERP resolution and are cached. Node features remain input-dependent and are interpolated on each forward pass. The same spherical coordinate convention is used for sampling and back-projection.

\subsection{SCC Operators and Initialization}
\label{app:scc_details}

The ERP and context inputs are separately normalized by GroupNorm. The affine predictor $f_{\mathrm{aff}}$ is a $1\times1$ convolution from $C$ to $2C$ channels, whose outputs are split into $S_0$ and $T$. We bound only the scale adjustment through $S=0.5\tanh(S_0)$.

The gate predictor processes the channel-wise concatenation $[\bar X,\bar G,\bar X-\bar G]$ using a $1\times1$ convolution from $3C$ to $C$ channels, GroupNorm, GELU, and a second $1\times1$ convolution with $C$ output channels. Applying sigmoid gives $M\in(0,1)^{B\times C\times H\times W}$.

The local branch applies a $1\times1$ projection from $C$ to $2C$ channels, GroupNorm and GELU, a $3\times3$ depth-wise convolution, GroupNorm and GELU, and a final $1\times1$ projection back to $C$ channels. The depth-wise convolution uses horizontal circular padding and vertical replicate padding. The context branch applies a $1\times1$ convolution, GroupNorm, GELU, and another $1\times1$ convolution, preserving $C$ channels throughout. The sum of the local and gated context updates is normalized by GroupNorm before residual addition. All GroupNorm operators use eight groups for the channel configurations in this model.

We initialize the affine predictor's weights and biases to zero, so $S=T=0$ and $X_c=X$ at initialization. This identity applies to the affine modulation stage. The final gate convolution has bias initialized to $-2$ to bias the gate toward smaller values at the start of training.

\subsection{Module Configurations}
\label{app:module_config}

\begin{table}[h]
\centering
\caption{Configurations of the spherical modeling modules at different feature scales.}
\label{tab:fsg_module_config}
\begin{tabular}{lccccc}
\toprule
Scale & $C$ & $N$ & $k$ & Graph Blocks & $k_b$ \\
\midrule
$1/4$  & 64  & 2584 & 16 & 2 & 4 \\
$1/8$  & 128 & 1597 & 12 & 2 & 4 \\
$1/16$ & 256 & 610  & 8  & 2 & 2 \\
$1/32$ & 512 & 233  & 8  & 2 & 2 \\
\bottomrule
\end{tabular}
\end{table}

Table~\ref{tab:fsg_module_config} summarizes the configurations of FSG and SCC at different feature scales. Here, $C$ denotes the feature dimension, $N$ the number of Fibonacci nodes, $k$ the neighborhood size for each spherical relation, and $k_b$ the number of neighbors used for spherical-to-ERP back-projection. The node count is resolution-dependent rather than manually fixed across scales. For each ERP feature resolution, we first determine the corresponding node budget and select the nearest Fibonacci number as $N$, resulting in progressively fewer nodes from fine to coarse feature scales. Each scale constructs independent local, dilated, and antipodal relations with its own learnable parameters, while two graph blocks are used consistently across all scales. The original UniFuse $1/2$-scale fusion branch is retained without FSG or SCC. Each cubemap face has a side length of $H_{\mathrm{in}}/2$, where $H_{\mathrm{in}}$ denotes the input ERP height.

\end{document}

%% file: math_commands.tex
\usepackage{amsmath,amsfonts,bm}

\def\eqref#1{equation~\ref{#1}}

\def\1{\bm{1}}

\DeclareMathAlphabet{\mathsfit}{\encodingdefault}{\sfdefault}{m}{sl}
\SetMathAlphabet{\mathsfit}{bold}{\encodingdefault}{\sfdefault}{bx}{n}

